\documentclass[letterpaper, 10 pt, conference]{ieeeconf}  % Comment this line out if you need a4paper

\IEEEoverridecommandlockouts                              % This command is only needed if 
\usepackage{multirow}
\usepackage{tikz}
\usepackage{pgfplots}
\usepackage{graphicx}
\usepackage{caption}
\usepackage{subcaption}
\usepackage{tabulary}

\usetikzlibrary{positioning}
\usetikzlibrary{arrows}

\title{\LARGE \bf
A Systematic Literature Review of Experiments in Socially Assistive Robotics using Humanoid Robots
}

\author{Floris Erich$^{1}$, Masakazu Hirokawa$^{2}$ and Kenji Suzuki$^{3}$% <-this % stops a space
\thanks{$^{1}$Floris Erich is with the School of Integrative and Global Majors (SIGMA), University of Tsukuba, 1-1-1 Tennodai, Tsukuba, Ibaraki, Japan.
        {\tt\small erich@ai.iit.tsukuba.ac.jp}}%
\thanks{$^{2}$Masakazu Hirokawa is with the Faculty of Engineering, Information and Systems, University of Tsukuba.
        {\tt\small hirokawa\_m@ieee.org}}%
\thanks{$^{3}$Kenji Suzuki is with the Faculty of Engineering, Information and Systems, University of Tsukuba.
        {\tt\small kenji@ieee.org}}%
}

\begin{document}

\maketitle
\thispagestyle{empty}
\pagestyle{empty}

%%%%%%%%%%%%%%%%%%%%%%%%%%%%%%%%%%%%%%%%%%%%%%%%%%%%%%%%%%%%%%%%%%%%%%%%%%%%%%%%
\begin{abstract}

We perform a Systematic Literature Review to discover how Humanoid robots are being applied in Socially Assistive Robotics experiments. Our search returned 24 papers, from which 16 were included for closer analysis. To do this analysis we used a conceptual framework inspired by Behavior-based Robotics. We were interested in finding out which robot was used (most use the robot NAO), what the goals of the application were (teaching, assisting, playing, instructing), how the robot was controlled (manually in most of the experiments), what kind of behaviors the robot exhibited (reacting to touch, pointing at body parts, singing a song, dancing, among others), what kind of actuators the robot used (always motors, sometimes speakers, hardly ever any other type of actuator) and what kind of sensors the robot used (in many studies the robot did not use any sensors at all, in others the robot frequently used camera and/or microphone). The results of this study can be used for designing software frameworks targeting Humanoid Socially Assistive Robotics, especially in the context of Software Product Line Engineering projects.

\end{abstract}

%%%%%%%%%%%%%%%%%%%%%%%%%%%%%%%%%%%%%%%%%%%%%%%%%%%%%%%%%%%%%%%%%%%%%%%%%%%%%%%%
% TO DO:
%  - Explicitly answer the research questions

\section{Introduction}
In this paper we present a conceptual framework for the study of robotic applications and our findings in applying the framework to the field of Humanoid Socially Assistive Robotics. 
% What is SAR?
Socially Assistive Robots are robots which exhibit social behavior for assisting a person in need. The field is on the intersection of assistive robotics and socially interactive / intelligent robotics \cite{Rabbitt15}. Humanoid Robots are robots which look like or behave like humans.

% Why Humanoid SAR?
% Some research about why the humanoid form factor is good for SAR would be nice here

% What is Behavior-based Robotics?
% Why Behavior-based Robotics?
The Behavior-based Robotics paradigm allows the construction of robots exhibiting complex operations based on the combination of simple behaviors \cite{Arkin98}. We believe that by making it easy for an end-user to combine these building blocks we can make robotics programming more accessible for inexperienced software developers \cite{Berenz14}.

The approach for enabling the end-user to combine these building blocks together is based on Software Product Line Engineering (SPLE) for Robotics \cite{Gherardi13}.
SPLE divides the process of developing software into two phases: Domain Engineering and Application Engineering \cite{Pohl2004}. 
The goal of Domain Engineering is to create models of a domain in which it is beneficial to reuse a significant portion of behavior. The goal of Application Engineering is to use the models created during the Domain Engineering phase to develop working applications.

The main question is how Humanoid robots are being applied in Socially Assistive Robotics experiments. We split this up into the following subquestions:
\begin{enumerate}
\item{What robot was developed or used?}
\item{What is/are the goal(s) of the robot application?}
\item{How was the robot controlled?}
\item{What kind of behaviors did the robot exhibit?}
\item{Which type of sensors were used?}
\item{Which type of actuators were used?}
\end{enumerate}

We found the third subquestion, how the robot was controlled, to be the hardest to answer. We considered the autonomy of the human subjects and of the robot. Experiments which have real human subjects and a robot which is controlled manually but covertly are called Wizard of Oz experiments. Various other combinations of ``Wizard'' and ``Oz'' are suggested to reflect on various levels of autonomy of the human subjects and of the robot \cite{Steinfeld09}. Experiments which are performed with real human subjects and a completely autonomous robot can be considered as ``Wizard \emph{and} Oz''. Our research is about \emph{applications} of robotics, so studies which do not have real human subjects have been excluded, as we do not consider these to be applications. Autonomy of a robot is a spectrum ranging from completely autonomous to completely manual/teleoperated. We classify studies based on three levels of this spectrum: Autonomous, mixed and manual.

In this paper we first introduce the Systematic Literature Review (SLR) methodology in Section \ref{Methodology}. Then in Section \ref{Results} we discuss the two major results of this research, which are a conceptual framework and information gathered by applying the conceptual framework. In Section \ref{Discussion} we then provide discussion of the results as well as address the concerns regarding validity and completeness of the applied research methodology. Finally we conclude the paper in Section \ref{Conclusion}.

\section{Methodology}
\label{Methodology}
% Explain SLR
We performed a Systematic Literature Review (SLR) \cite{Kitchenham07}, which is a methodology inspired from the medical sciences. It allows for an exhaustive synthesis of the literature regarding a certain topic to be performed. Our goal for performing this SLR was to create a complete overview of the types of applications developed in the field of Humanoid Socially Assistive Robotics.

To select studies for inclusion in this review we applied the search term \emph{(social or socially) and assistive and (robots or robotics) and humanoid} on the \emph{Web of Science} database. The composite search string allows us to find a compact yet complete list of papers indexed by the database searched. This search was last performed on the 1st of February 2016 and returned 24 papers. We used one inclusion criterium, that to be included a paper should discuss a primary study of the implementation of a robotic application and should hence not be limited to a theoretical treatise of robotic application construction or be a review article. We used these exclusion criteria: (1) Paper is written in a language other than English and (2) paper is inaccessible through the databases of our university. 

To decide whether the paper met the inclusion criteria and did not meet any of the exclusion criteria we read at least the title and abstract for every paper returned by the search. All the papers included in the study were completely read, and we extracted data from them using the conceptual framework.

\section{Results}
\label{Results}

The results of this study are a summary of the literature, a conceptual framework which can be used to extract data from the literature and the information gathered using the conceptual framework.

\subsection{Summary of the literature}

Before we describe the robotic application in-depth, we would first like to give a high level overview of the studies which we included in this review. The robots used in the experiments are: 
\begin{enumerate}
\item KASPAR: A child-sized humanoid robot developed by the University of Hertfordshire. 
\item NAO: A child-sized research robot made by Aldebaran Robotics. Its Choreographe programming interface makes it suitable to design complex movement choreographies and scripted plays by novice programmers. 
\item Bandit: A robot with a humanoid upper body and a wheel base. 
\item Robota: A robot doll which can move its legs, head and arms. 
\item Robovie R3: A child-sized humanoid robot with a wheel base, developed by Vstone in collaboration with the Advanced Telecommunications Research Institute.
\end{enumerate}

Robots can be used to train children with Autism Spectrum Disorder (ASD). Two studies which try to accomplish this use the robot KASPAR. In both studies the robot is used for training tactile interaction. In the first of these studies \cite{Costa15}, KASPAR points at various body parts (e.g. head, nose, tummy) and asks the child to do the same, first as separate events, then in sequence, and finally by singing and dancing. In the second study using KASPAR \cite{Robins10} the robot reacts to different kinds of touches, depending on the location and the strength (gentle or rough). In another study Robota has been used to evaluate the response of children with ASD \cite{Robins06}. The reaction of the children is compared with the reaction to an actor pretending to be a robot.

Robots can be used as an educational instrument. In one study we encountered NAO assisting staff in kindergartens by going through a nine phase procedure (introducing itself, singing a song, initiating personal contact, explaining its limitations regarding conversations, falling down, explaining its limitations regarding mobility and finally parting ways) \cite{Fridin14a}. In another study the robot performed a play, and the authors surveyed teaching staff's acceptance of humanoid robots \cite{Fridin14b}. NAO has also been used to teach children about exercises to prevent lower back pain \cite{Magyar15}.

Robots have potential to be used to facilitate elderly care. NAO was embedded in an Intelligent Home, in which it did not only interface with a elderly person but also various external systems such as a domotic (home automation) platform \cite{Torta14}. This was actually the most advanced application we encountered in this review.

Another application of robots is by offering friendly support to a patient undergoing invasive treatments. NAO was used in a study which considered the impact of a robot in assisting children undergoing cancer treatment \cite{Alemi14}. The robot acts like it is also suffering from illness. Through various sessions the children get acquainted with the robot. The robot does pretend play to inform the child of various aspects of the disease and how it affects the child's life style.

While most studies we encountered studied the response from humans to the robot on a macro level (i.e.\ the robot is used in reasonably long sessions, sometimes divided over a few days, and the robot's presence in an environment is measured), in some cases a robot is used to study more technical characteristics of Human Robot Interaction, taking a micro level perspective (short sessions, humans reactions to subtle behaviors of the robot are measured). In one micro level study a NAO robot hands a letter over to a participant, sometimes doing this smoothly, but sometimes resisting \cite{Baddoura14, Baddoura15}.

NAO has also been used for studying interaction with children who have ASD. In one experiment the robot is combined with a camera network in order to make the robot able to adapt its behavior according to the cues of a child's head movement \cite{Bekele13}. In another study two models of interaction are proposed, either interaction of the child with just a robot or interaction of the child with a robot and an adult \cite{Zhao13}. In this latter experiment the robot application is expository, as the goal of the application is to measure how children react to a robot in general.

Robots can also be used to instruct people about dangers or about general information. In a study performed in a hotel \cite{Pan15}, two scenario's use a single NAO and one scenario uses two NAO's. In the first two scenario's the NAO greets hotel guests and instructs the guests about hotel information. In the dual-NAO scenario two robots have a conversation about the hotel.

One experiment uses Bandit to assist the recovery of a post-stroke patient \cite{Wade11}. The robot interfaces with a wire puzzle to accomplish this.

NAO and Robovie have also been used in an experiment to teach children Turkish Sign Language \cite{Uluer15,Kose15}. NAO turns out to be less effective in this scenario due to its limited amount of fingers. The modified version of Robovie R3 used in the research has five fingers and is hence more suitable for this type of research.
\subsection{Conceptual Framework}

While doing the research we iteratively constructed a conceptual framework to store the knowledge gained during the study. If any concept would get added or removed to the framework we would verify whether this influenced the data extraction from any of the previous papers studied. This approach is based on the Grounded Theory methodology of performing qualitative research \cite{Glaser67}. If this research would be repeated by a different researcher (while using the same methodology), the conceptual framework constructed should be similar.

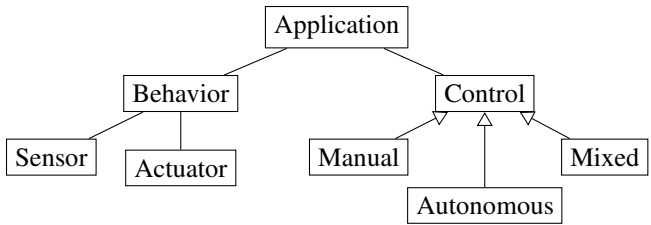
\begin{figure}[t!]
\vspace*{0.15cm}
\centering
\begin{tikzpicture}
\begin{scope}[yshift=8cm,node distance=2cm and 1cm]

\node[draw] (Application) at (0,0) {Application};

\node[draw, below left = 0.5 cm of Application](Behavior){Behavior};

\node[draw,below = 0.5 cm of Behavior](Actuator){Actuator};

\node[draw,below left = 0.5 cm of Behavior](Sensor){Sensor};

\node[draw, below right = 0.5cm of Application] (Control) {Control};

\node[draw,below left = 0.5 cm of Control](Manual){Manual};

\node[draw,below = 1 cm of Control](Autonomous){Autonomous};

\node[draw,below right = 0.5 cm of Control](Mixed){Mixed};

\draw[-] (Application) -> (Behavior);

\draw[-] (Behavior) -> (Actuator);

\draw[-] (Behavior) -> (Sensor);

\draw[-] (Application) -> (Control);

\draw[open triangle 60-] (Control) -> (Manual);

\draw[open triangle 60-] (Control) -> (Mixed);

\draw[open triangle 60-] (Control) -> (Autonomous);
\end{scope}
\end{tikzpicture}
\caption{Behavior-based Conceptual Framework of SAR applications. Regular lines represent association. The open triangle arrow represents an is-a relationship.}
\label{fig:conceptualframework}
\end{figure}

Figure \ref{fig:conceptualframework} shows the conceptual framework we are using to evaluate the studies. The conceptual framework defines the following concepts: 

\begin{itemize}
\item Robot: Each study used one or more robots for which the an application was developed. 
\item Goal: Each robotic application tries to accomplish one or more goals. 
\item Behavior: To achieve the goals the robot has to exhibit one or more behaviors. 
\item Actuator: To exhibit a behavior a robot has to use one or more actuators, which are hardware components which can make changes to the world. In this research we are interested in the type of actuator. 
\item Sensor: Sensors are hardware components which retrieve information from the environment in which the robot is deployed. We only concern ourselves with sensors providing exteroception, i.e.\ perception of things happening outside of the robot embodiment \cite{Arkin98}. 
\item Control: Robotic applications define behaviors in terms of combinations of sensors, behaviors and actuators. Control concerns how components of these three types are connected. We define three types of control: Autonomous, mixed or manual (Wizard of Oz). 
\item Autonomous Control: Control exercised by some automated system, such as algorithms running on the robot. Any control action taken by the robot in response to an action performed by a research subject is also considered autonomous. 
\item Manual Control: Control by a human operator - typically the experimenter - through a programming interface of the robot. 
\item Mixed Control: Mixed control by both some automated system and a human operator.
\end{itemize}

Metadata for the studies also includes the title, authors, year of publication and research goals. 

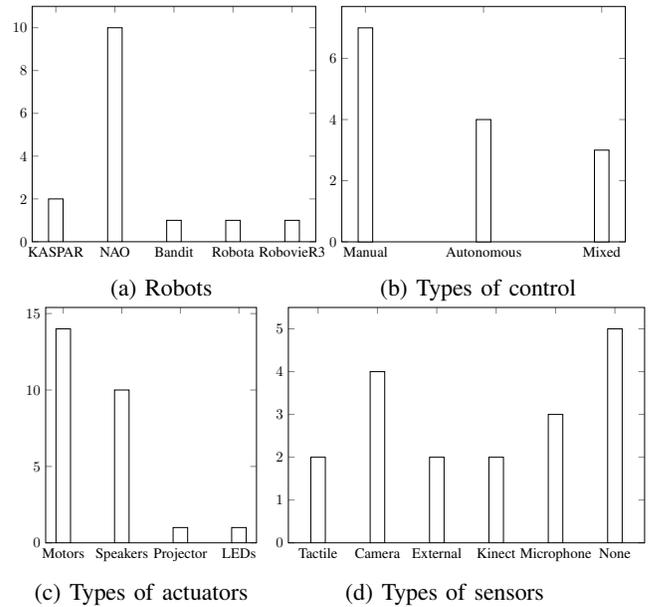
\begin{figure}[t!]
\vspace*{0.15cm}
\centering
\begin{subfigure}[b]{0.23\textwidth}
    \begin{tikzpicture}[scale=0.55]
        \begin{axis}[
            symbolic x coords={KASPAR, NAO, Bandit, Robota, RobovieR3},
            ymin=0,
            xtick=data
          ]
            \addplot[ybar,fill=white] coordinates {
                (KASPAR,   2)
                (NAO,  10)
                (Bandit, 1)
                (Robota, 1)
                (RobovieR3,   1)
            };
        \end{axis}
    \end{tikzpicture}
    \caption{Robots}
\end{subfigure}
\begin{subfigure}[b]{0.23\textwidth}
    \begin{tikzpicture}[scale=0.55]
        \begin{axis}[
            symbolic x coords={Manual, Autonomous, Mixed},
            ymin=0,
            xtick=data
          ]
            \addplot[ybar,fill=white] coordinates {
                (Manual,   7)
                (Autonomous,  4)
                (Mixed,   3)
            };
        \end{axis}
    \end{tikzpicture}
    \caption{Types of control}
\end{subfigure}
\begin{subfigure}[b]{0.18\textwidth}
    \begin{tikzpicture}[scale=0.55]
        \begin{axis}[
            symbolic x coords={Motors, Speakers, Projector, LEDs},
            ymin=0,
            xtick=data,
            width=190pt,
            height=\axisdefaultheight
          ]
            \addplot[ybar,fill=white] coordinates {
                (Motors,   14)
                (Speakers,  10)
                (Projector, 1)
                (LEDs,   1)
            };
        \end{axis}
    \end{tikzpicture}
    \caption{Types of actuators}
\end{subfigure}
\begin{subfigure}[b]{0.26\textwidth}
    \begin{tikzpicture}[scale=0.55]
        \begin{axis}[
            symbolic x coords={Tactile, Camera, External, Kinect, Microphone, None},
            ymin=0,
            xtick=data,
            width=290pt,
            height=\axisdefaultheight
          ]
            \addplot[ybar,fill=white] coordinates {
                (Tactile,   2)
                (Camera, 4)
                (Microphone,  3)
                (Kinect,   2)
                (External, 2)
                (None, 5)
            };
        \end{axis}
    \end{tikzpicture}
    \caption{Types of sensors}
\end{subfigure}
\caption{Quantitative evaluation of the studies.}
\label{fig:quantitativeresults}
\end{figure}

\subsection{Information gathered}

From the 24 studies returned by the search, we selected 16 studies which met the inclusion and exclusion criteria. The full bibliography is available online\footnote{URL: http://floriserich.nl/wordpress/humanoid-socially-assistive-robotics-slr/}. The excluded papers and the reason for exclusion can also be found online.

% Excluded:
    % Excluded, only concerns the construction of a robot, not its actual application
    % McGin, 2014    & Towards the Design of a New Humanoid Robot for Domestic Applications \\
    % Excluded, theoretical treatise
    % Simou, Artificial Humanoid for the Elderly People
    % Excluded, not a primary study
    % Hersh, 2015    & Overcoming Barriers and Increasing Independence - Service Robots for Elderly and Disabled People \\
    % Excluded, not a primary study
    % Soares, 2013   & Robotica-Autismo Project: Technology for autistic children \\
    % Excluded, does not concern a robotics application
    % Wu, 2012       & Designing robots for the elderly: Appearance issue and beyond \\
    % Excluded, does not concern an robotics application but rather technology which can be used in robotic applications
    % Carcagni, 2014 & Visual Interaction Including Biometrics Information for a Socially Assistive Robotic Platform \\
    % Excluded, not an interactive robot, more like a static doll...
    % Mead, 2013     & Automated Proxemic Feature Extraction and Behavior Recognition: Applications in Human-Robot Interaction \\
    % Excluded, not available
    % Dinet, EXPLORATORY INVESTIGATION OF ATTITUDES TOWARDS ASSISTIVE ROBOTS FOR FUTURE USERS

\begin{table*}[t]
  \vspace*{0.1cm}
  \centering
  \caption{Information gathered from the studies.}
  \begin{tabular}{ 
  p{0.5cm}             p{1.1cm}    p{3.5cm}          p{1.15cm}  p{4.5cm} p{2cm} p{1.8cm}
  }
  \hline
  Ref.        & Robot(s)  & Application Goal & Control & Behaviors & Actuators & Sensors \\
  \hline
  % key:
  \cite{Costa15} & 
  % robot:
  KASPAR &
  % goal:
  Teach children with ASD to identify their body parts and increase their body awareness & 
  % control:
  Mixed & 
  % behaviors:
  Reacting to touch; identifying body part and asking participant to match;
  identifying sequence of body parts and asking the participant to match; and
  singing a song while dancing and encouraging the participant to join &  
  % actuators:
  Motors, speakers
  &
  % sensors: 
  Tactile
  \\
  % key:
  \cite{Robins10} &
  % robot:
  KASPAR &
  % goal: 
  Teach children with ASD about various topics related to their self, their body and social interaction & 
  % control:
  Autonomous &
  % behaviors:
  Respond to touch in different areas by moving the body of the robot & 
  % actuators:
  Motors &
  % sensors:
  Tactile 
  \\
  \cite{Fridin14a} &
  % robot:
  NAO &
  % goal: 
  Assist staff in kindergartens &
  % controller:
  Manual &
  % behaviors:
  Singing a song while dancing, initiating personal contact, playing a game, falling/getting up, giving explanations &
  % actuators:
  Motors, speakers & 
  % sensors:
  Microphone 
  \\ % check if a microphone was actually used
  \cite{Torta14} & 
  % robot:
  NAO &
  % goal:
  Assisting in elderly care & 
  % controller:
  Autonomous & 
  % behaviors:
  Providing environmental information; playing music; managing phone calls; monitoring self treatment; monitoring the environment; providing video calls & 
  % actuators:
  Motors, speakers, projector
  &
  % sensors:
  Microphone, camera, external sensor network
  \\
  % key:
  \cite{Alemi14} &
  % robot:
  NAO &
  % goal:
  Play a role-taking game with a patient &
  % control:
  Manual & 
  % behaviors:
  Pretend play using talking, gesturing and playing music & 
  % actuators:
  Motors, speakers & 
  % sensors:
  None used
  \\
  % key:
  \cite{Fridin14b} & 
  % robot:
  NAO &
  % goal:
  Interact with teachers &
  % control:
  Mixed &
  % behaviors:
  Detecting nearby people, talking to people, grasping &
  % actuators:
  Motors, speakers &
  % sensors:
  Camera 
  \\
  % key:
  \cite{Baddoura14, Baddoura15} &
  % robot:
  NAO &
  % goal:
  Deliver a letter &
  % control:
  Manual &
  % behaviors:
  Walking, bowing, handing over a letter, waving &
  % actuators:
  Motors &
  % sensors:
  None used
  \\
  % key:
  \cite{Bekele13} &
  % robot:
  NAO &
  % goal: 
  Test and train children with ASD about attention skills &
  % control:
  Mixed &
  % behaviors:
  Asking questions while moving naturally &
  % actuators:
  Motors, speakers &
  % sensors:
  Camera network
  \\
  % key:
  \cite{Zhao13} &
  % robot:
  NAO &
  % goal: 
  Interacting with an ASD child &
  % control:
  Manual &
  % behaviors:
  Sitting, moving / dancing, speaking &
  % actuators:
  Motors, speakers &
  % sensors:
  None used 
  \\
  % key:
  \cite{Pan15} &
  % robot:
  NAO &
  % goal: 
  Engage with and instruct hotel guests &
  % control:
  Autonomous &
  % behaviors:
  Looking at hotel guests, reading from a script &
  % actuators:
  Motors, speakers (Text-To-Speech) &
  % sensors:
  Kinect
  \\
  % key:
  \cite{Wade11} &
  % robot:
  Bandit &
  % goal: 
  Assist individuals post-stroke &
  % control:
  Autonomous &
  % behaviors:
  Giving instructions / feedback / motivating, pointing, nodding &
  % actuators:
  Motors, speakers &
  % sensors:
  Wire puzzle
  \\
  % key:
  \cite{Robins06} &
  % robot:
  Robota &
  % goal: 
  Interact with an autistic child &
  % control:
  Manual &
  % behaviors:
  Move according to operator's instructions &
  % actuators:
  Motors &
  % sensors:
  None used
  \\
  % key:
  \cite{Uluer15, Kose15} & 
  % robot:
  Robovie R3, NAO &
  % goal: 
  Tutor sign language to a child &
  % control:
  Mixed &
  % behaviors:
  Indicating signs from Turkish Sign Language &
  % actuators:
  Motors, LED's, speakers &
  % sensors:
  Kinect, camera, microphone
  \\
  % key:
  \cite{Magyar15} &
  % robot:
  NAO &
  % goal: 
  Teach exercises to prevent back pain &
  % control:
  Manual &
  % behaviors:
  Demonstrate exercises to subjects &
  % actuators:
  Motors &
  % sensors:
  None used
  \\
  \hline
  \end{tabular}
  \label{table:review}
\end{table*}

The information which we gathered is summarized per study in Table \ref{table:review}. The studies are categorized by the robot used for executing the robotic application. The application goal specifically refers to the goal of the robot application and not the goals of the research project in which the robot is being applied. Control can be either manual, autonomous or a mix of both. The behavior column should be considered from the perspective of the robot. Also, when we list sensors and actuators we only mention components interacting with the environment outside of the robot, instead of any internal sensing or actuation which might occur. Figure \ref{fig:quantitativeresults} gives an overview of the quantified results.

\section{Discussion}
\label{Discussion}

We will first discuss the results for each question and finally discuss some of the limitations of our research methodology.

\subsection{Robot}

NAO is by far the most used robot in experiments for Humanoid SAR. Most papers however did not discuss the choice of which robot to use in detail. One study shows that for some experiments NAO is limited due to its hand not being similar to a human hand in terms of the amount of digits \cite{Uluer15, Kose15}. This is an issue if the robot has to make detailed hand gestures, such as when using Sign Language.

One of the studies compared a \emph{faux} robot\footnote{We did not include the \emph{faux} robot in our results.} (played by a mime artist) with an actual robot \cite{Robins06}, because the authors argued that there were no robots available at the time to accurately replicate human facial expressions. Today robotics has advanced to a point in which robots are actually able to do this (for example Geminoid), but these type of robots are probably still out of economic reach for most studies.

\subsection{Application goal}

The applications vary between highly generic and highly specific applications. We believe this is caused by a difference in the type of studies, some validating the design of a robotic application and some trying to study some psychological or social construct. Studies of the former type concern the question of whether a robot can be used for a certain type of application (\emph{Design Science} questions \cite{Wieringa14}), while studies of the latter type ask questions about more basic constructs, such as how people react to a robot in a certain environment (basic science questions). Both questions are worth asking when trying to design a new robotic application, and it can be argued that to answer the Design Science questions one has to use basic science questions.

\subsection{Control}

One of the findings which surprised us was the low interactivity provided by the robot in most studies. Even though most studies used versatile robots such as NAO, the control of the robot was often manual, while few sensors were being used. One study mentioned that manual control was beneficial in these kind of experiments because of favoring repeatability of the experiment \cite{Baddoura14}. This argument seems to only hold for research which asks basic science questions, as manual control by an experimenter could be perceived as a limitation for studies focusing more on Design Science questions. The usage of teleoperated robots in experiments is also called Wizard of Oz.

It is normal for research projects involving robotics to use a controlled environment, which also means that the experimenter is tightly involved in the operations of the robot. However, we believe that to measure the real impact robotics can have in society we it is necessary to apply the robot in an environment matching its target environment as closely as possible, which implies that there is no experimenter involved in operating the robot. We were however impressed that four studies were already using fully autonomous robots. 

\subsection{Behavior}

In all of the studies the behavior exhibited by the robot was partly scripted. A typical design pattern is to manually start up the scripted activities, though in some cases the robot would automatically start the scripted activities after receiving an environmental stimuli. We could describe these stimuli as being discrete, e.g.\ \emph{guest present} and \emph{sensor touched}. The high prevalence of scripted behavior is discouraging from the perspective of behavior-based robotics, which tries to construct tight feedback loops between a robot and the environment, in which stimuli are continuous and the reaction is often in proportion to the stimuli. Some studies considered the use of interactive behaviors to be future work, suggesting techniques from the fields of computer vision and machine learning as the means of realizing more interactive behavior.

\subsection{Actuation}

In terms of actuation the findings were quite homogeneous, with all of the robots using motors and most of them using speakers. We excluded one paper in which the robot did not perform any sensing and actuation at all \cite{Mead13}. %It could be argued that the response generated by motors and speakers correspond with a human's primary means of communication, i.e.\ body language and verbal communication. 

\subsection{Sensing}

Some studies considered the sensing capabilities of the robots they used to be too limited. While NAO is equipped with two cameras for example, the limited resolution of each make it unsuitable for performing complex computer vision tasks. Most studies ended up not using any sensors at all, but we are not sure whether this was because of limited sensing capabilities of the robot, limited processing capabilities of the robot, limited programming skills of the experiment designers, perceived lack of control over the experiment or some other reason. Some studies show that sensing capabilities of a robot can be extended by using external sensor systems such as the Kinect controller and entire sensor networks. 

% Application goals are sometimes simplistic as the robot application are often written as a means to an end. 
% One interesting thing is that Baddoura mentioned feedforward control is good for reproducability of research. It is surprising that so few applications use a fully autonomous robot. This is discouraging from the perspective of behavioral-based robotics, but might signal that a hybrid approach of control would be beneficial. This validates research into Robot Runtime Adoption. 
% Most applications seem to be quite low fidelity, mostly using simply motors and speakers. Even the usage of external sensors is quite limited. This is partly because of the limited capabilities of the robot, e.g. Bekele mentions how limited Nao's cameras are. This might also be related to limiting the research to robots with humanoid shape, which might be more constraining for application designers.

\subsection{Towards higher autonomy in robotic applications}

Humanoid Socially Assistive Robotics is arguably a complex field. The locomotion of a legged humanoid robot is more sophisticated than wheeled mobile robots \cite{Siegwart11}. Experiments in this field often require review by Institutional Review Boards. The robots are expected to perform tasks ordinarily performed by a human, or at least supplement these in a significant way. Perhaps these factors contribute to the limited amount of autonomy which can be attributed to most robotic applications we evaluated.

There are software solutions which make it easy for a robot to perform complicated locomotion. One example is Choreographe from Aldebaran, which was also frequently used in the experiments which we reviewed. This software makes it easy for robots to act in a scripted way, and is quite accessible for novices. At the same time there are various behavior-based architectures which allow robots to perform in a dynamic way. We developed the TDM (Targets-Drives-Means) architecture for controlling robots \cite{Berenz14}. 

TDM applications consist of a set of \emph{behaviors} which a robot can perform (such as walk towards an object or greet a person), for each behavior a \emph{schema} which detects an object in the real world (such as a human or a toy) which activates the behaviors, and a set of \emph{score calculators} (such as distance to the object) which determine which behaviors take precedence when there are multiple schemes active \cite{Berenz11}. Compared to flowchart based approaches, TDM takes over the management of state and sequence. Experiments using the TDM architecture show that it is more difficult to begin with than traditional approaches such as flowcharts, but is considered to be easier to manipulate at later stages \cite{Berenz12}, i.e.\ there is a shorter learning curve. Further experimentation with the architecture show that for end-users it is beneficial to use a graphical interface for writing applications \cite{Berenz14}.

\subsection{Limitations}

Regarding the methodology used to perform the review, we believe the validity of the findings is guaranteed by having two reviewers discuss the materials selected for inclusion. For the sake of validity it should be disclosed that one of the studies (by Pan et al.) included in our review was written by a member of our laboratory. We did not discuss any of our findings with the primary author of that study, however both studies were supervised by the principal investigator of our laboratory. Another important issue with the validity of our findings is that we relied on the primary studies for accurately reporting the gathered information. If for example the authors used some actuator such as LED's but did not find it worthwhile to mention this in the paper it will also be missing from our results.

Our review included all papers returned by the search term applied to Web of Science, so is in that sense complete. However, to increase completeness further more databases can be added (such as Scopus) and constraints can be removed from the search term (for example by removing the keyword humanoid\footnote{Search term: \emph{(social or socially) and assistive and (robots or robotics)}}: 237 results). Furthermore the research could be extended by performing a forward-backward citation analysis, which would respectively consider the papers citing the included studies and the papers cited by the included studies. We consider this research to be a prototype for applying this methodology and have hence decided to focus on this constrained search term and single database.

\section{Conclusion}
\label{Conclusion}

The contribution of this research is two-fold. First, it sketches a methodology for studying previous robotic applications as reported in academic literature, which can be used during the Domain Engineering phase of a Robotics Software Product Line Engineering process. Second, it presents an overview of applications for Humanoid Socially Assistive Robots, which is of general interest, and will be used by ourselves as a basis for a software framework.

\subsection{Future work}
We are expanding the set of studies by doing a forward/backward citation analysis. One round of forward analysis has already been performed, leading to eight more studies to be included. The latest work in progress version of this research can be found online\footnote{URL: http://floriserich.nl/wordpress/humanoid-socially-assistive-robotics-slr/}.

We propose that robot programming can be supported using a graphical interface and that this interface requires a notation adapted to the specific domain. In this research we have explored the domain of Humanoid Socially Assistive Robotics with the goal of discovering the type of applications being developed, so we can develop a specialized version of TDM for this domain. This version should contain the necessary programming elements used within this domain, as well as a graphical notation to support application development by novices, rooted in sound principles of notation design \cite{Moody09}.

We will incorporate the findings into a domain specific version of a robotics programming platform. The applications using autonomous control are especially interesting as they show the kind of applications for which our platform could be applied. For the other applications we should evaluate how they can be improved to reach autonomous control.

Our way of analysis of the studies can be considered as \emph{black box}, as we did not have direct access to the study results and relied on study reports. We want to additionally approach the application of Humanoid SAR from a \emph{white box} perspective, i.e.\ with full access to the data collected during the studies. We want to test whether it is possible to extract behavioral scripts by logging the experiments and applying \emph{process mining} techniques.

\bibliographystyle{IEEEtran}
\bibliography{biblio}

\end{document}